\documentclass[3p,twocolumn]{elsarticle}

\usepackage{hyperref}
\usepackage{subfigure}
\journal{Journal of Artificial Intelligence in Medicine}

\newcommand{\TempCoNet}{LWFNet}
\newcommand{\GRUNet}{\TempCoNet}
\newcommand{\NaiveNet}{Naive\GRUNet}

\begin{document}

\begin{frontmatter}

\title{Unsupervised temporal context learning using convolutional neural networks for laparoscopic workflow analysis}
\author[IAR]{Sebastian Bodenstedt}\corref{cor1}
\ead{bodenstedt@kit.edu}
\author[HD]{Martin Wagner}
\author[IAR]{Darko Katic}
\author[HD]{Patrick Mietkowski}
\author[HD]{Benjamin Mayer}
\author[HD]{Hannes Kenngott}
\author[HD]{Beat M\"uller-Stich}
\author[IAR]{R\"udiger Dillmann}
\author[IAR]{Stefanie Speidel}
\cortext[cor1]{Corresponding author}
\address[IAR]{Institute for Anthropomatics and Robotics, Karlsruhe Institute of Technology, Karlsruhe}
\address[HD]{Department of General, Visceral and Transplant Surgery, University of Heidelberg, Heidelberg}
\begin{abstract}
Computer-assisted surgery (CAS) aims to provide the surgeon with the right type of assistance at the right moment.
Such assistance systems are especially relevant in laparoscopic surgery, where CAS can alleviate some of the drawbacks that surgeons incur.
For many assistance functions, e.g. displaying the location of a tumor at the appropriate time or suggesting what instruments to prepare next, analyzing the surgical workflow is a prerequisite.
Since laparoscopic interventions are performed via endoscope, the video signal is an obvious sensor modality to rely on for workflow analysis.

Image-based workflow analysis tasks in laparoscopy, such as phase recognition, skill assessment, video indexing or automatic annotation, require a temporal distinction between video frames.
Generally computer vision based methods that generalize from previously seen data are used.
For training such methods, large amounts of annotated data are necessary.
Annotating surgical data requires expert knowledge, therefore collecting a sufficient amount of data is difficult, time-consuming and not always feasible.

In this paper, we address this problem by presenting an unsupervised method for training a convolutional neural network (CNN) to differentiate between laparoscopic video frames on a temporal basis.
We extract video frames at regular intervals from 324 unlabeled laparoscopic interventions, resulting in a dataset of approximately 2.2 million images.
From this dataset, we extract image pairs from the same video and train a CNN to determine their temporal order.
To solve this problem, the CNN has to extract features that are relevant for comprehending laparoscopic workflow.

Furthermore, we demonstrate that such a CNN can be adapted for surgical workflow segmentation.
We performed image-based workflow segmentation on a publicly available dataset of 7 cholecystectomies and 9 colorectal interventions.
\end{abstract}


\end{frontmatter}

%\linenumbers

\section{Introduction}
The aim of a computer-assisted surgery system (CAS) is to provide the surgeon with the right type of assistance at the right moment. 
In laparoscopic surgery, such a system could be used to compensate for some of the drawbacks typical to laparoscopy, such as the limited field of view or difficult orientation in the abdominal cavitiy, by e.g. providing assistance during navigation.

For many applications in CAS, such as providing the position of a tumor, specifying the most probable tool required next by the surgeon or determining the remaining duration of surgery, analyzing the surgical workflow is a prerequisite.
Since laparoscopic surgeries are performed using an endoscopic camera, a video stream is always available during surgery, making it the obvious choice as input sensor data for workflow analysis.
Many workflow analysis tasks, e.g. phase recognition, skill assessment, automatic reporting, video indexing or automatic annotation, require a method for providing a temporal representation of video frames, or rather their content.

Often, laparoscopic tool usage \cite{Bouarfa2011455}\cite{Stauder2014}\cite{Padoy2012632} or surgical activities \cite{Katic2014}\cite{Neumuth2006}\cite{Forestier2015} are used as feature for such a representation, but currently this information is usually derived through additional hardware (e.g. RFID tags in the case of \cite{Stauder2014}), which is not generally available in the OR or through manual annotation, which is not feasible for online workflow segmentation or large datasets.
The kinematic data from a robotic system, such as the daVinci can be used for providing tool usage information and tool trajectories \cite{DiPietro}\cite{Zappella2013732}, but this information is only available for robotic interventions and not the majority of laparoscopic interventions.

While methods for automatically extracting information on tool usages from endoscopic images do exist \cite{Blum2010}\cite{Speidel09} there are few publications with a purely image-based approach for workflow analysis \cite{Blum2010}\cite{Dergachyova2016}\cite{Lalys2012}\cite{TwinandaSMMMP16}\cite{Lea2016}.
The authors in \cite{Blum2010}, \cite{Dergachyova2016} and \cite{Lalys2012} utilize a combination of manually selected image features to describe the content of single video frames.
Manually selecting image features has the drawback that only information that the domain expert is aware of can be captured, other characteristics that might still contribute are possibly lost. 

In computer vision, one possible solution to the issue of feature selection are convolutional neural networks (CNN), a type of artificial neural network, which has the ability to learn image features. CNNs are currently the state of the art in many areas in computer vision, such as object detection and image classification \cite{girshick2014rich} \cite{Alexnet}.

In \cite{TwinandaSMMMP16}, the authors propose EndoNet, a combination of a CNN and a hybrid hidden markov model (HHMM).
The CNN here is used to automatically learn image features that can be used to distinguish different surgical phases in laparoscopic gallbladder removals, which are then fed into a HHMM to determine the most probable phase for each image frame.
On the dataset of the Endoscopic Vision 2015 Workflow Challenge\footnote{\label{EndoVis}http://endovissub-workflow.grand-challenge.org/} (EndoVis15Workflow), EndoNet outperforms the method outlined in \cite{Dergachyova2016}, which uses manually selected image features.
The drawback of EndoNet is that a large amount of annotated data is used for training, 40 videos of laparoscopic gallbladder removals in which not only the surgical phases, but also the laparoscopic instruments are annotated for each frame.
This amount of annotated data is difficult and costly to collect.
If one takes into consideration that laparoscopic gallbladder removals are simple and standardized operations, one can assume that more complex types of interventions, such as colorectal or pancreatic surgery, would require even more labeled data.
In \cite{Lea2016}, the authors present a CNN-based approach for offline phase detection that outperforms EndoNet on the EndoVis15Workflow dataset, which uses only 6 operations for training.
Offline phase detection means that data from the entire intervention is used for assigning a phase to each frame retrospectively.
The approaches makes usage of spatio-temporal information to capture object motion during the course of a laparoscopic intervention.
The features extracted with the CNN are then combined with either a linear model, a semi-markov model or a time-invariant model, based on dynamic time warping, with the latter two models outperforming \cite{TwinandaSMMMP16}, leading to the conclusion that including temporal information during workflow analysis improves classification outcome.

One of the advantages of CNNs is that it is possible to take a CNN that is solving one task (e.g. detecting cars) and retrain it for solving a different task (e.g. detecting bicycles) \cite{erhan2010does}.
Retraining (or pretraining), instead of training a new CNN from scratch, has the advantage that previously learned features (say features that respond to wheels) can be repurposed.
By repurposing features, a pretrained CNN should require less training data to achieve adequate performance.
Generally training deep CNNs requires a large amount of annotated data, which, especially in a surgical environment, is not always feasible to obtain, since usually experts are required to annotate data.
Pretraining the CNN using unlabeled data would therefore be preferable.

In \cite{doersch2015unsupervised}, the authors train a CNN to develop an understanding of the spatial context of different excerpts from a given image.
For this, they divide unlabed images into multiple 3x3 box grids and train a CNN to arranged the outer blocks correctly in relation to the center block.
Part of this trained CNN is then modified and retrained to partake in an object detection challenge, achieving state of the art results. 
Inspired by \cite{doersch2015unsupervised}, we extended the idea of pretraining a CNN with spatial context information to pretraining with temporal context information provided by given videos.

In this paper, we propose a method for a CNN to learn visual features by sorting frames from videos of laparoscopic interventions into the correct temporal order.
We assume that the features learned while solving the sorting task enable the CNN to distinguish frames based on their temporal context.
Such a CNN can be used as starting point for many applications were these visual cues would be beneficial, e.g. online and offline video segmentation, automatic annotation, indexing and generating surgical reports.
This temporal context learning task is performed using unlabeled laparoscopic videos.

Furthermore, we evaluate the suitability of such a pretraining for a supervised workflow segmentation task, in this case, segmenting surgical videos into phases (surgical phase detection).
For this, the pretrained CNN is extended to take information from the current frame and previous frames to deduce the phase of the current frame.
In contrast to other approaches, this method does not rely on manually-selected features or expensive annotation of surgical tools and, furthermore, is able to detect phases online.
The evaluation is performed on two datasets, the EndoVis15Workflow dataset, which contains 7 annotated cholecystectomies and a dataset containing 9 laparoscopic colorectal interventions, a more complex type of surgery, recorded in the University Hospital of Heidelberg.

\section{Unsupervised temporal context learning}
\label{sec:tcl}
In this section, we present our method for training such a deep CNN using unlabeled videos.
We accomplish this by solving a task that requires the CNN to sort two given frames into the correct temporal order.
For this, a large dataset from multiple laparoscopic interventions is used.
We assume that solving such a task requires the CNN to learn to extract visual cues that describe the temporal flow of laparoscopic interventions.

\subsection{Training task}
\begin{figure}[tb]
\centering
   \includegraphics[width=1.0\columnwidth]{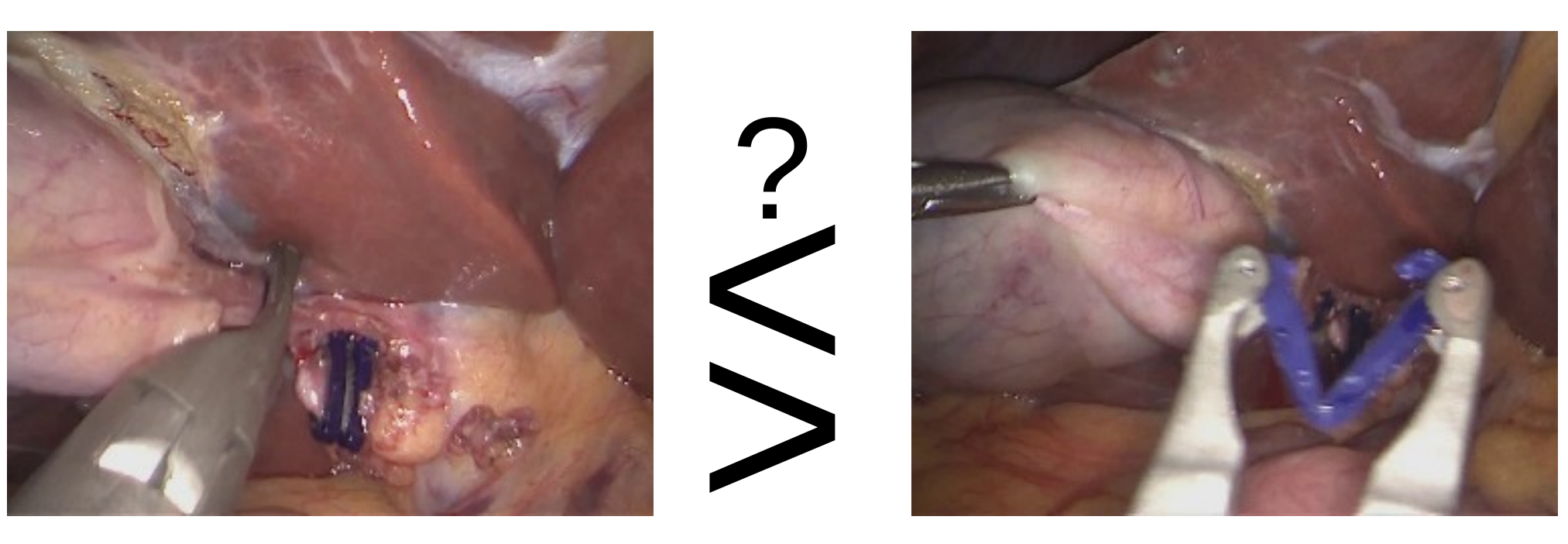}
   \caption{Our task for pretraining a CNN. 
   Which is the most probable temporal order of the two images? 
   (Answer: the right image comes first, the clip has to be inserted into the body, before being placed.)}
   \label{fig:order}       % Give a unique label
\end{figure}
The task we propose for training the CNN is illustrated in figure \ref{fig:order}: Given two frames from the same laparoscopy, what is the most probable relative order of the two frames, i.e. which frame comes first?
We uniformly sample two random frames from the video of a laparoscopic intervention and feed it into our CNN.
The CNN must then compute the relative order of the two frames in the original video, i.e. which frame comes first.
We assume that solving this task requires the CNN to extract visual cues relevant to surgical workflow and thereby develop an understanding of the temporal flow of laparoscopic interventions.
\subsection{Dataset}
\label{sec:dataset}
To train the CNN, we used a large dataset consisting of 324 laparoscopic interventions recorded anonymously at the University Hospital of Heidelberg.
The dataset contains videos of 30 different types of laparoscopic interventions, providing a diverse range in training data. 
The videos were all recorded in the same operating room using the integrated operating room system OR1\texttrademark\ (Karl Storz GmbH \& Co KG, Tuttlingen, Germany).
The interventions were performed by multiple surgeons with varying endoscopes and optics.
We extracted frames at intervals of one frame per second, resulting in approximately 2.2 million images.
Since the videos were recorded automatically, we had to ensure that sequences that did not contain any large changes (e.g. black screens) were excluded from the dataset.
This was accomplished by excluding a video frame $f$ from the dataset, if for the last video frame $g$ from the same video that was included in the dataset $$||I(f)-I(g)|| < 8000$$, with $I(f)$ and $I(g)$ being the respective pixel values for each image.
\subsection{Training the CNN}
\begin{figure}[tb]
\centering
   \includegraphics[width=1.0\columnwidth]{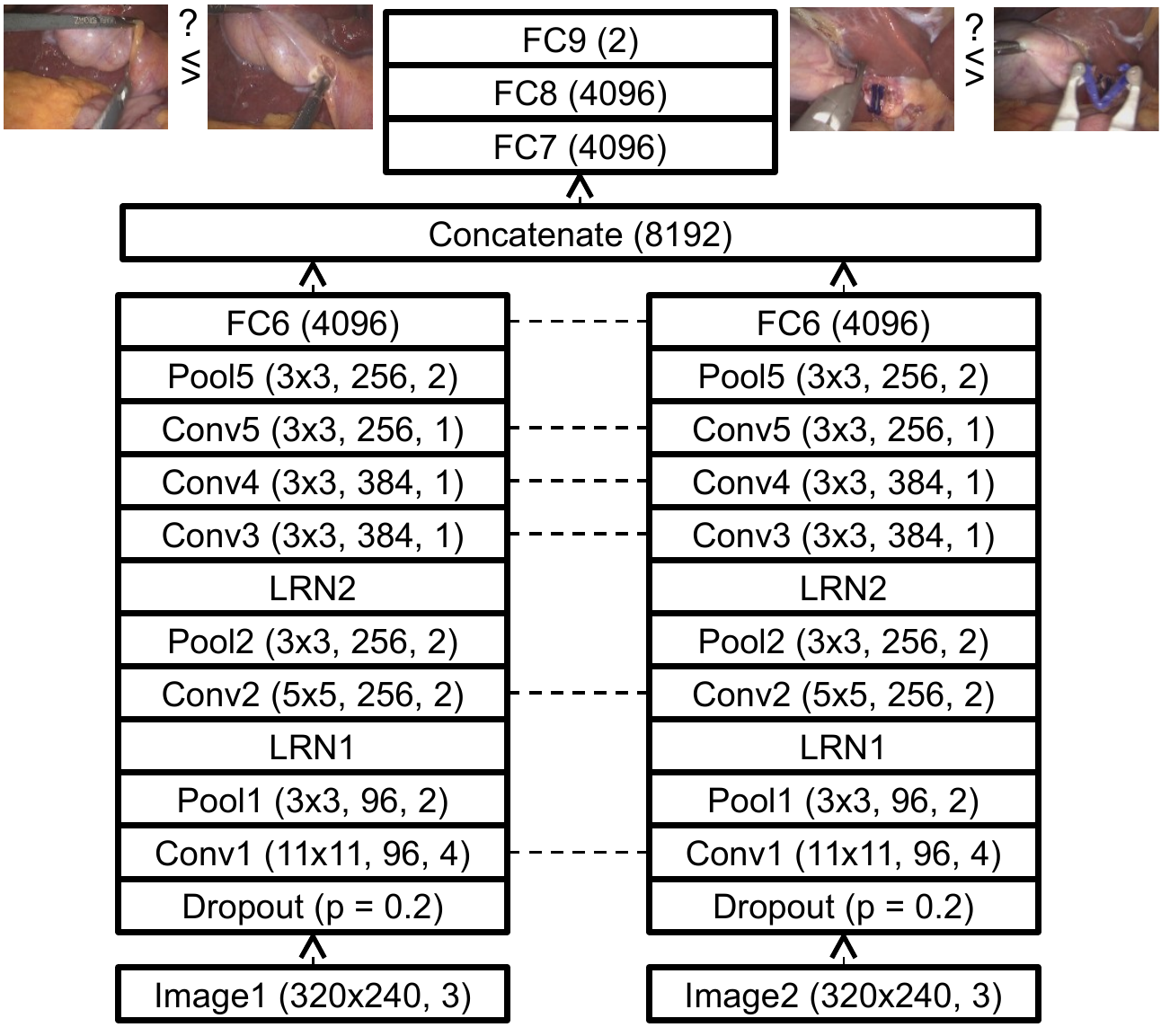}
   \caption{CNN Topology for the temporal context prediction task: Dotted lines indicate shared weights. 
	\textit{Dropout} are dropout layers that, with a probability of $p$, set a value to zero. 
	\textit{Conv} are convolutional layers, \textit{LRN} are local response normalization layers\cite{Alexnet}, \textit{Pool} are max-pooling layers, \textit{FC} are fully connected layers and \textit{Concatenate} concatenate two input vectors. 
	The numbers in parenthesis indicate size of filter kernel, number of outputs and step size. 
	In the case of fully connected layers, the number of hidden units is listed instead.}
   \label{fig:tcl}       % Give a unique label
\end{figure}

Selecting a network topology that allows a CNN to predict the relative order of two given video frames from scratch can be a difficult task.
We therefore based our model on the one presented in \cite{doersch2015unsupervised}, which was shown to work for spatial context prediction.
The topology of the network used can be seen in fig. \ref{fig:tcl}.
A pair of frames from the same video is fed into the two input layers of the CNN.
Each frame is then processed by a chain of multiple convolutional layers (Conv1 to Conv5), each with AlexNet-style topology \cite{Alexnet}, resulting in a reduced representation of the frame in a fully connected layer (FC6).
The corresponding layers in both chains share weights.
The outputs of the two FC6 layers are then concatenated and then processed using two further fully connected layer.
FC9 then outputs if either frame 1 (Output: 0) oder frame 2 (Output: 1) comes first in a temporal order.
For every convolutional and fully connected layer, except FC8, a ReLu (rectified linear unit) nonlinearity\cite{icml2010_NairH10}  was used.
FC9 uses a softmax nonlinearity instead.

During training, for each epoch (iteration) we sample with replacement 256 operations out of all operations.
From each of these operations, 3 frames, $I_1$, $I_2$ and $I_3$, are drawn randomly, with $I_t < I_{t+1}$ or, in other words, $I_t$ precedes $I_{t+1}$ in a temporal order.
The frames are then resampled to a resolution of $320\times240$.
To ascertain that the proportions inside the frames are not skewed by this, we crop the borders of the images to give the image a $4:3$ aspect ratio in case they exhibited a different ratio.
Furthermore, we normalize each value in the RGB channels by mapping them into the range of $[-0.5,0.5]$.
We then form 6 inequations, i.e.
$$ I_0 < I_1,\ I_0 < I_2,\ I_1 < I_2 $$
$$ I_1 > I_0,\ I_2 > I_0,\ I_2 > I_1 $$
resulting in 1536 inequations per epoch.
The CNN is then trained for 10000 epochs using stochastic gradient descent (learning rate of 0.0005) combined with nesterov momentum (momentum of 0.9).
As loss function, we selected categorical cross-entropy.
The CNN was implemented in Python, using Theano\cite{Theano} and Lasagne\cite{Lasagne}, and trained using NVidia GTX Titan X and NVidia GTX 1080.
\section{Laparoscopic workflow segmentation}
For a given laparoscopic frame, the method outlined in section \ref{sec:tcl} provides a descriptor that makes a temporal distinction possible.
In this section, we determine the suitability of such a descriptor for surgical workflow segmentation, i.e. dividing a given surgical in coherent and semantic meaningful segments.
\subsection{\textit{\NaiveNet}}
\begin{figure}[tb]
\centering
   \includegraphics[width=1.0\columnwidth]{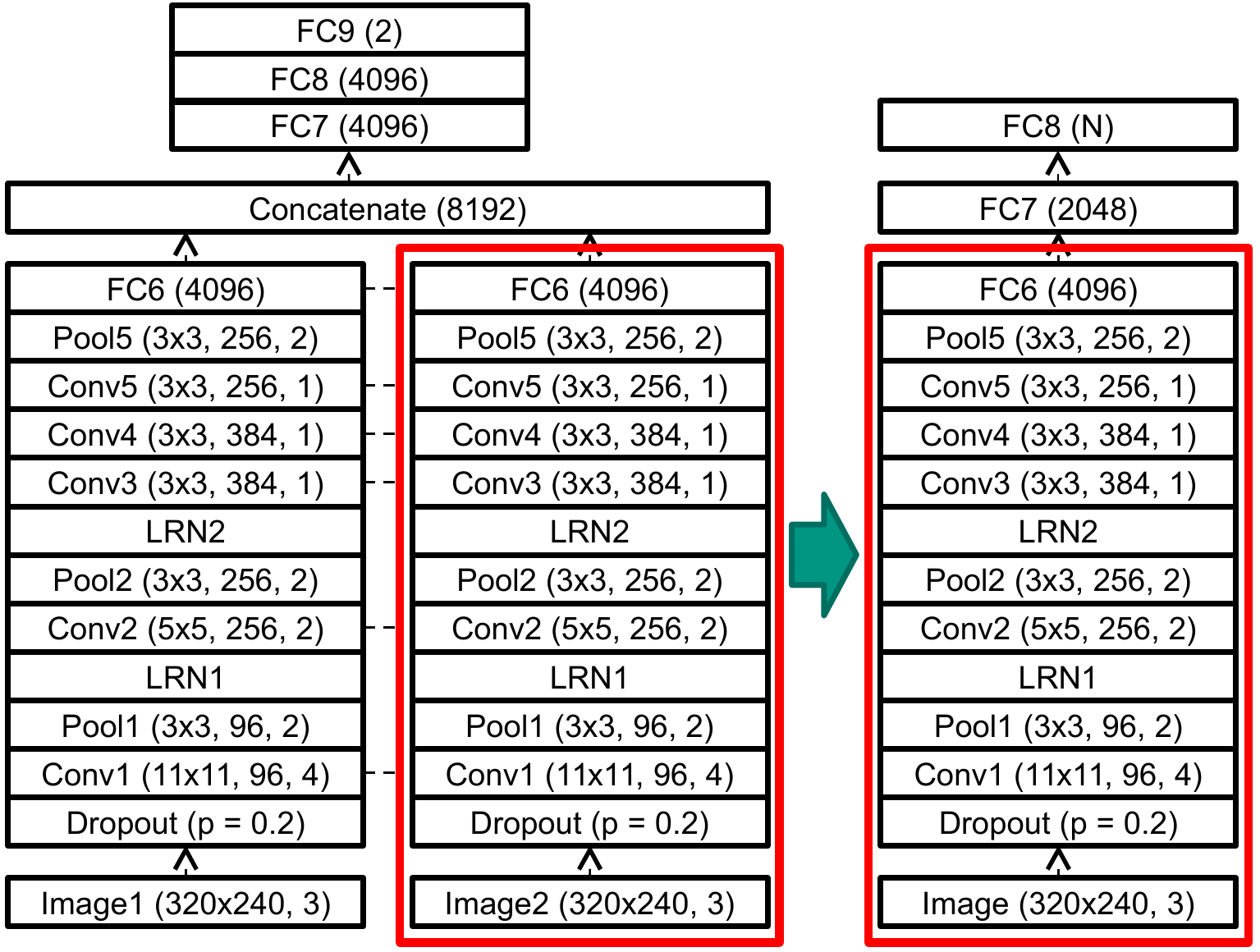}
   \caption{\textit{\NaiveNet}: For a naive approach to laparoscopic phase detection, we take part of the CNN illustrated in fig. \ref{fig:tcl} and add two further fully-connected layers to assign a class. Here $N$ indicates the number of phases.}
   \label{fig:phase}       % Give a unique label
\end{figure}

A naive approach to workflow segmentation would be to extend one of the processing chains (everything before FC6) with further fully-connected layers to assign each frame to the most probable class label.
We constructed a naive CNN for laparoscopic workflow analysis (\textit{\NaiveNet}, Naive \textbf{L}aparoscopic \textbf{W}ork\textbf{F}low \textbf{Net}work) as can be seen in fig. \ref{fig:phase}.

While distinguishing frames certainly is a prerequisites for laparoscopic phase detection, determining the current state from just a single frame seems questionable and prone to ambiguities.
We assume that single frames alone do not contain sufficient information to deduce the current phase and therefore propose to extent \textit{\NaiveNet} to include information seen in previous frames.
\subsection{\textit{\GRUNet}}
\begin{figure}[tb]
\centering
   \includegraphics[width=0.7\columnwidth]{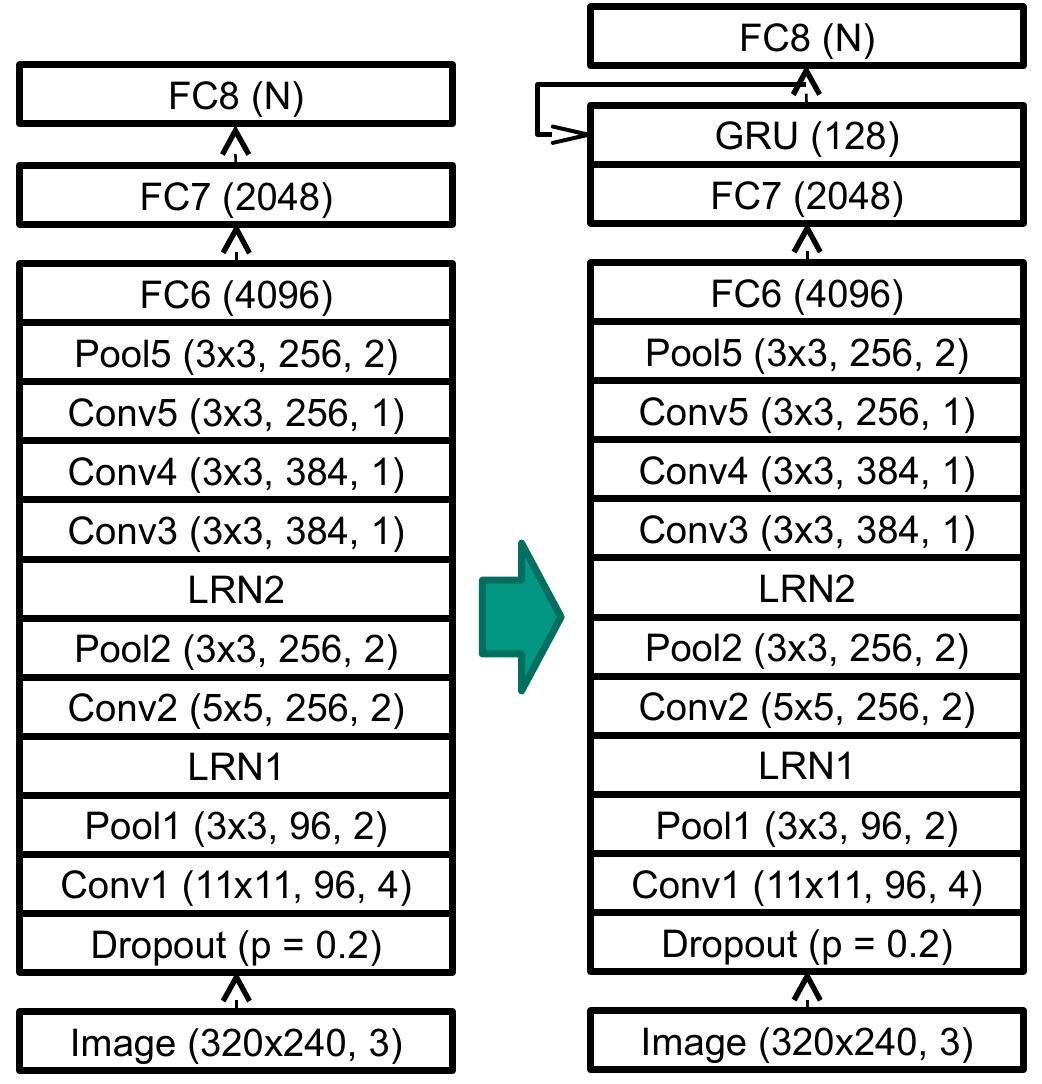}
   \caption{\textit{\GRUNet}: To incorporate previously seen information into our approach for workflow segmentation, we \textit{\NaiveNet} and combine it with a gated recurrent unit (GRU)\cite{ChoMBB14}, which makes it possible to retain information from previous frames.}
   \label{fig:phaselstm}       % Give a unique label
\end{figure}

Feedforward neural networks, by definition, do not contain cycles and therefore do not recollect previous states to compute the current output.
Recurrent neural networks (RNN) overcome this limitation by introducing cycles in the topology of the network and thereby allowing the network to process sequences.
Tradition RNNs suffer from multiple drawbacks, such as gradients that vanish over the course of training and recalling only ``recent'' information \cite{hochreiter1997long}.
Long term-short term memory units (LSTM)\cite{hochreiter1997long}, a deep RNN architecture, do not suffer from these drawbacks and, furthermore, are selective about the information they retain and forget.
Similar to LSTMs, gated recurrent units (GRU)\cite{ChoMBB14} also do not suffer from the drawbacks of traditional RNN architecture and can learn to recall/forget particular information.
Seeing as GRUs perform similarly to LSTMs for certain tasks\cite{chung2014empirical}, whilst having fewer parameters, we decided to extend \textit{\NaiveNet} with a GRU (fig. \ref{fig:phaselstm}) into \textit{\GRUNet}.
To integrate the GRU into \textit{\GRUNet}, the output from FC6 has to be modified slightly, as RNNs expect sequences as input.
For this, we reshape the output from FC6, a 2D tensor of the shape $batch size\times 4096$, to a 3D tensor of shape $1 \times batch size\times 4096$, simulating a $batch size$ long sequence. 
Generally, the number of frames in a video exceeds the batch size, meaning that, instead of one long sequence, the GRU only sees multiple shorter sequences.
To compensate for this, we take the contents of the hidden state after the last element of the sequence and use it to initialize the hidden state before processing the next batch.
\subsection{Training}
The CNN is trained using stochastic gradient descent (initial learning rate $\lambda_0$ was set to $10^{-3}$) combined with nesterov momentum (momentum of 0.9) for multiple epochs with a batch size of 256.
To penalize large weights and thereby prevent overfitting, we apply L1 and L2 regularization during training.
For this, we add terms to the cost function, which incorporate the L1 and L2 norm of the weights and thereby penalize large weights, lowering the risk of overfitting.
We selected a weight of $10^{-5}$ for the L1 penalty term and $10^{-3}$ for the L2 penalty term.
To ensure convergence, we reduced the learning rate $\lambda$ by factor $\alpha$: $\lambda_{t+1} = \alpha\cdot\lambda_t$.
For $\alpha$, we selected 0.975 as value.
Since we are only interested in fine-tuning the parameters learned in section \ref{sec:tcl}, we use a smaller learn rate $\lambda'_t = 10^{-1}\cdot\lambda_t$ for FC6 and all layers proceeding it.
The value for the parameters specified here were determined empirically.
\section{Evaluation}
We evaluated the presented approaches for workflow segmentation on two datasets for laparoscopic phase detection.
To compare our proposed method to the state of the art, we first evaluate on the publicly available EndoVis15Workflow dataset.
Furthermore, to show that our method translates to longer, more complex interventions, we evaluate our method on a dataset comprised of colorectal interventions from the University Hospital of Heidelberg.
\subsection{Metrics}
The following metrics were used to evaluate the performance of the different workflow segmentation methods on a given video from a laparoscopic intervention:
\begin{itemize}
 \item Precision: Percentage of frames correctly attributed to a certain phase
 \item Recall: Percentage of frames attributed to a certain phase that are correctly attributed to that phase
 \item Accuracy: Overall percentage of frames attributed to the correct phase
\end{itemize}
For each analyzed video, we will compute the average over all phases for precision and recall.
\subsection{EndoVis15Workflow}
\begin{figure}[tb]
	\centering
	\subfigure[\textit{\NaiveNet}]{
	\label{NoLSTM}
	\includegraphics[width=1.0\columnwidth]{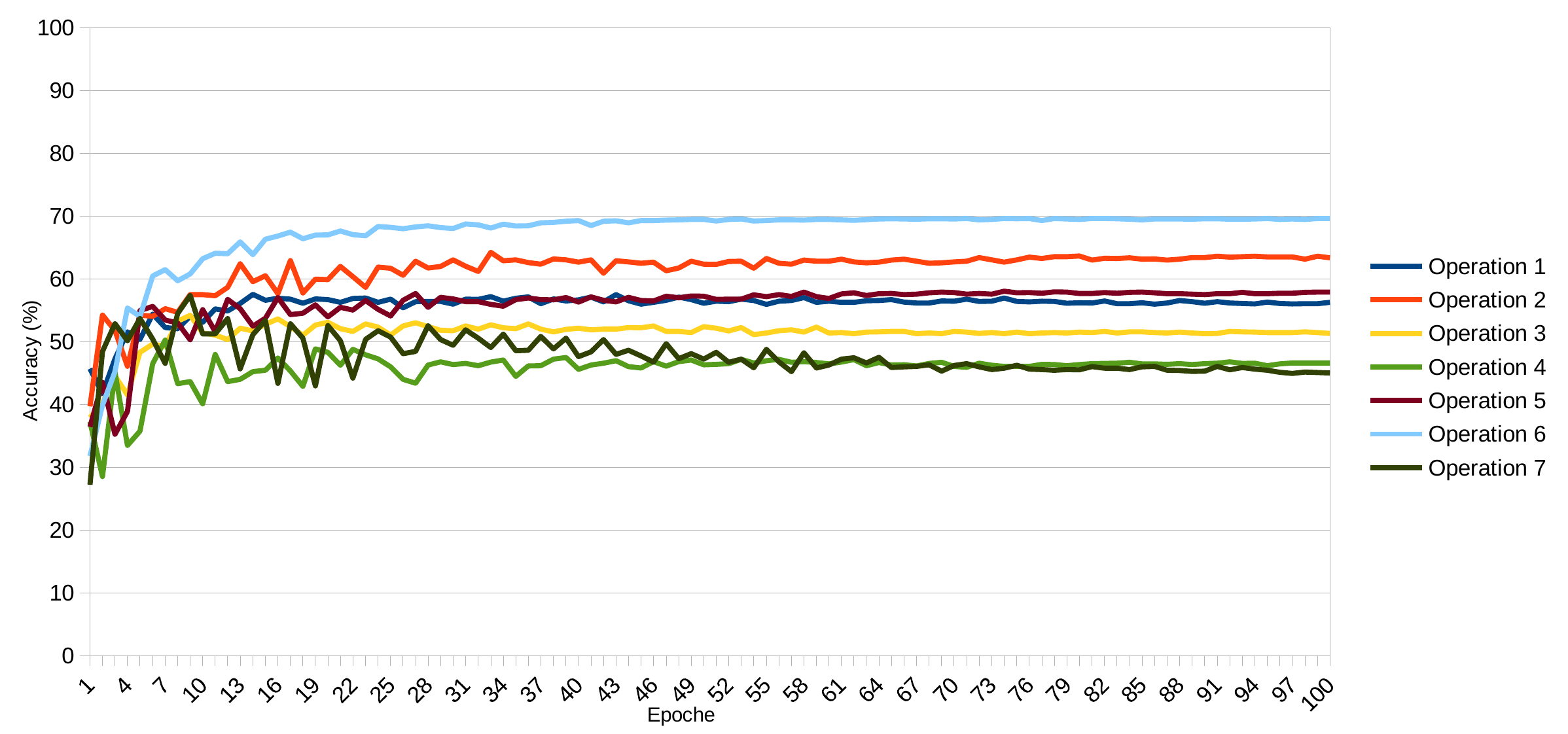}
	}
	\subfigure[\textit{\GRUNet}]{
	\label{LSTM}
	\includegraphics[width=1.0\columnwidth]{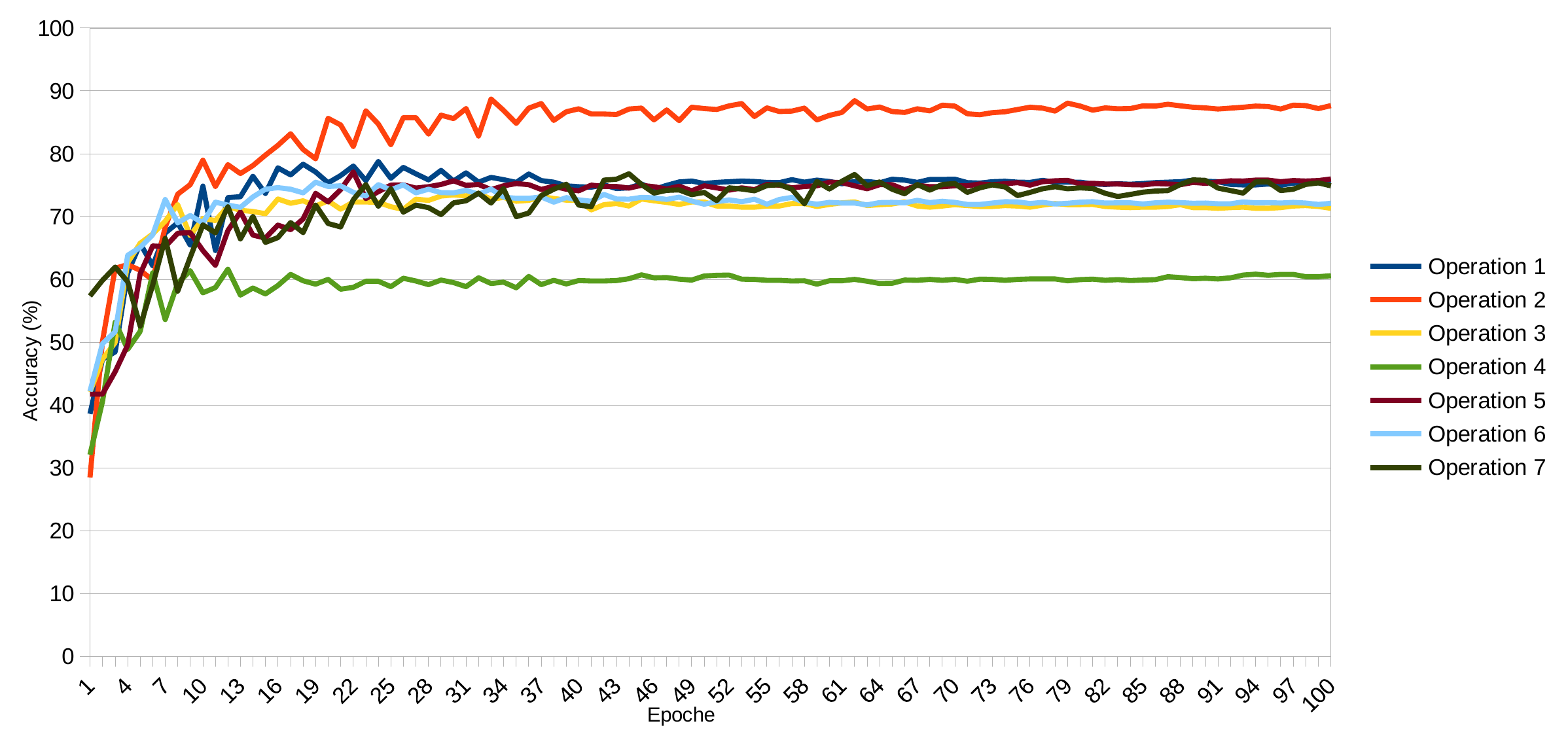}
	}
	\subfigure[\textit{\GRUNet} without pretraining]{
	\label{NoLSTMNoPretrain}
	\includegraphics[width=1.0\columnwidth]{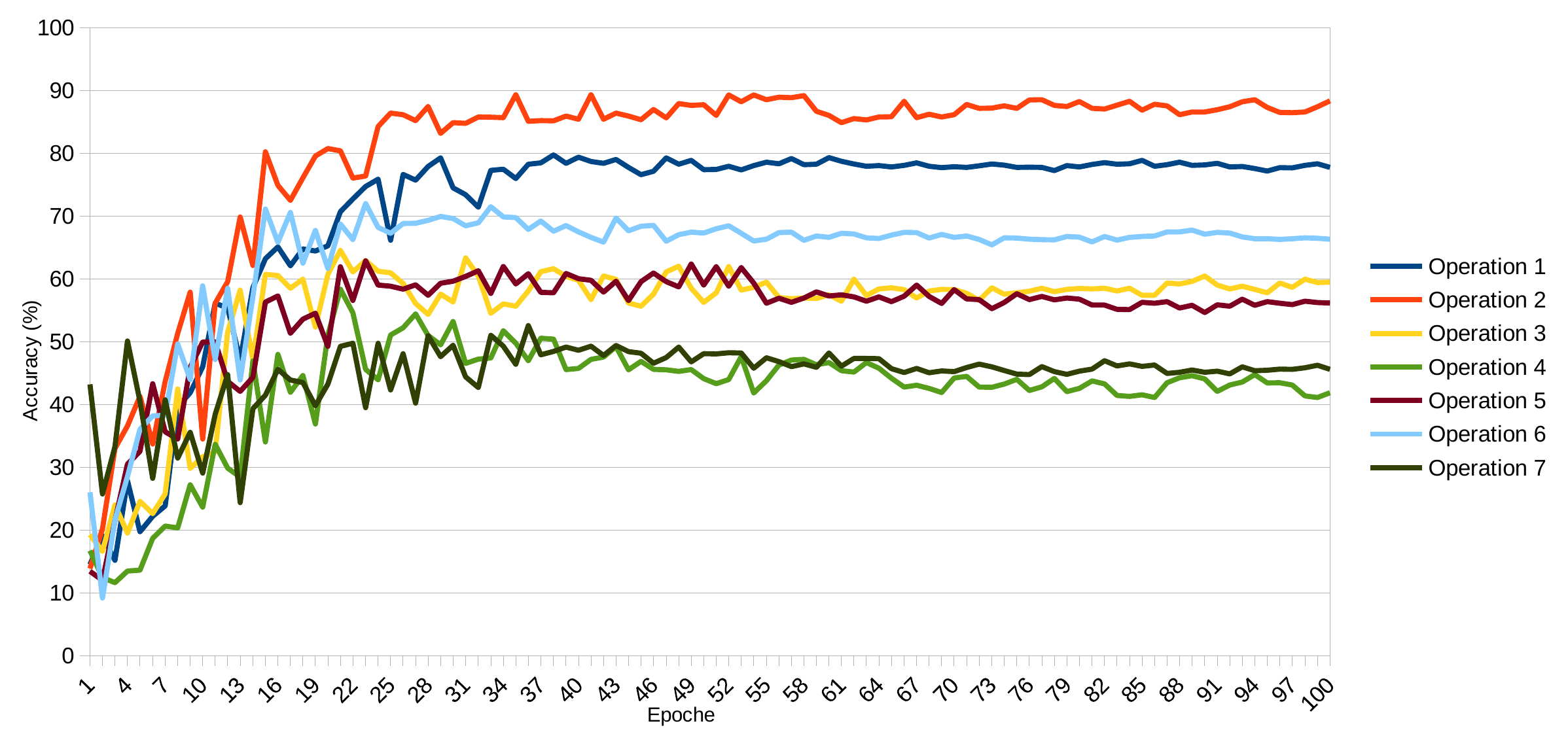}
	}
	\caption{Development of the accuracies of the phase detection for each operation and for each network during the leave-one-surgery-out evaluation on the EndoVis dataset.}
	\label{fig:gallphaselstm}
\end{figure}

The public dataset from the EndoVis 2015 workflow challenge consists of 7 laparoscopic cholecystectomies provided by the Technische Universit\"at M\"unchen.
The videos have been segmented into surgical phases, seven phases in total (tab. \ref{tab:phases}).
For each video frame the corresponding label was provided as annotation.
\begin{table}[htb]
\centering
\begin{tabular*}{\columnwidth}{l|l}
Phase ID & Explanation \\
\hline
1 &Placement of trocars \\
2 &Preparation of Calot’s triangle \\
3 &Clipping and cutting of cystic artery \\
  &and duct\\
4 &Gallbladder dissection \\
5 &Gallbladder retrieval \\
6 &Hemostasis \\
7 &Attaching drainage, wound closure\\
  &and end of operation \\
\hline
\end{tabular*}
\caption{Different phases in EndoVis15Workflow.}
\label{tab:phases}
\end{table}
To train both \textit{\NaiveNet} and \textit{\GRUNet}, we first sampled the provided videos at a rate of one frame per second, in order to reduce the data and thereby the time required for training.
We also resampled the resolution of the selected frames from $1920\times1080$ to $320\times240$.
Using this slightly modified data, we then performed a leave-one-surgery-out evaluation (training on 6 videos and testing on the 7\textsuperscript{th} video for all seven possible combination of training videos).
For each test set, we trained for 100 epochs.
The development of the accuracies for each run can be found in figure \ref{fig:gallphaselstm}.
To demonstrate the advantage of the proposed pretraining, we also included results for a version of \textit{\GRUNet} with randomly initialized weights in figure \ref{NoLSTMNoPretrain}.
Figure \ref{fig:gallphaselstm} clearly shows that the GRU based methods outperform the feedforward based \textit{\NaiveNet}.
Furthermore, we are also able to demonstrate that pretraining \textit{\GRUNet} as outlined in section \ref{sec:tcl} increases performance when compared to randomly initialized parameters.
Table \ref{tab:results} further highlights this, as it shows that \textit{\GRUNet} with pretraining achieves a higher precision, recall and accuracy in comparison to \textit{\GRUNet} without pretraining and \textit{\NaiveNet}.
\begin{table}[htb]
\centering
\resizebox{\columnwidth}{!}{
\begin{tabular}{|l|l|l|l|}
\hline
& Precision & Recall & Accuracy \\
\hline
\textit{\NaiveNet} & 56.6\% $\pm$ 7.5\%& 53.7\% $\pm$ 8.8\% & 56.3\%\ $\pm$ 8.1\%\\
\hline
\textit{\GRUNet} & 79.3\% $\pm$ 8.1\%& 73.7\% $\pm$ 9.7\% & 74.5\%\ $\pm$ 8.4	\%\\
\hline
\textit{\GRUNet} without pretraining & 75.4\% $\pm$ 11.8\%& 68.8\% $\pm$ 12.6\% & 66.0\%\ $\pm$ 14.8\%\\
\hline
\hline
EndoNet (CNN only)\cite{TwinandaSMMMP16} & 64.8\% $\pm$  7.3\%& 64.3\% $\pm$ 11.8\% & 65.9\% $\pm$ 4.7\%\\
\hline
EndoNet (CNN + HHMM)\cite{TwinandaSMMMP16} & 83.0\% $\pm$  12.5\%& 79.2\% $\pm$ 17.5\% & 76.3\% $\pm$ 5.1\%\\
\hline
Dergachyova et al.\cite{Dergachyova2016} & 72.1\% $\pm$ 16.4\%& 71.3\% $\pm$ 13.6\% & 68.1\%\\
\hline
\end{tabular}
}
\caption{Comparison of the results of our proposed methods, EndoNet \cite{TwinandaSMMMP16} (only online results) and the method proposed by Dergachyova et al.\cite{Dergachyova2016}.}
\label{tab:results}
\end{table}

We also compared our results to those published by Twinanda et al. \cite{TwinandaSMMMP16} and Dergachyova et al. \cite{Dergachyova2016} (tab. \ref{tab:results}).
\textit{\GRUNet} outperforms the method presented by Dergachyova et al. and the CNN only version of EndoNet.
The CNN + HHMM based EndoNet outperforms \textit{\GRUNet}, which can be attributed the large task specific dataset used for training EndoNet.

\begin{table}[hbt]
%\begin{table*}[t]
\centering
\resizebox{\columnwidth}{!}{
% \begin{tabular}{|l|l|l|l|l|l|l|l|}
% \hline
% & P1 & P2 & P3 & P4 & P5 & P6 & P7\\
% \hline
% Precision & 85.2\%$\pm$12.0\%&81.8\%$\pm$8.3\%&72.0\%$\pm$25.8\%&71.7\%$\pm$34.2\%&77.5\%$\pm$23.1\%&78.4\%$\pm$23.6\%&88.4\%$\pm$15.2\%\\
% \hline
% Recall & 98.3\%$\pm$4.4\%&89.0\%$\pm$9.6\%&64.0\%$\pm$34.0\%&55.8\%$\pm$41.6\%&83.3\%$\pm$14.1\%&51.6\%$\pm$37.9\%&73.9\%$\pm$26.7\%\\
% \hline
% Accuracy & 98.6\%$\pm$1.1\%&94.0\%$\pm$5.2\%&89.3\%$\pm$5.1\%&88.3\%$\pm$4.4\%&92.5\%$\pm$5.6\%&88.4\%$\pm$5.8\%&97.8\%$\pm$1.6\%\\
% \hline
% \end{tabular}

\begin{tabular}{|l|l|l|l|}
\hline
& Precision & Recall & Accuracy\\
\hline
P1 & 85.2\%$\pm$12.0\%&98.3\%$\pm$4.4\%&98.6\%$\pm$1.1\%\\
\hline
P2 & 81.8\%$\pm$8.3\%&89.0\%$\pm$9.6\%&94.0\%$\pm$5.2\%\\
\hline
P3 & 72.0\%$\pm$25.8\%&64.0\%$\pm$34.0\%&89.3\%$\pm$5.1\%\\
\hline
P4 & 71.7\%$\pm$34.2\%&55.8\%$\pm$41.6\%&88.3\%$\pm$4.4\%\\
\hline
P5 & 77.5\%$\pm$23.1\%&83.3\%$\pm$14.1\%&92.5\%$\pm$5.6\%\\
\hline
P6 & 78.4\%$\pm$23.6\%&51.6\%$\pm$37.9\%&88.4\%$\pm$5.8\%\\
\hline
P7 & 88.4\%$\pm$15.2\%&73.9\%$\pm$26.7\%&97.8\%$\pm$1.6\%\\
\hline
\end{tabular}
}
\caption{Performance of \textit{\GRUNet} broken down into the different phases.}
\label{tab:results2}
\end{table}

Table \ref{tab:results2} shows how \textit{\GRUNet} performs for each of the 7 phases individually.
The phases closes to the start and the finish achieve the highest performance in all metrics, while phase further away perform somewhat worse.
Of all phases, phase 6 has the lowest accuracy and recall, which can be attributed to the fact that phase 5 and 6 are often intermingled and visually very similar, making them difficult to distinguish.
Phase 4 also has a low performance, which could be explained by mix-ups with phases 3 and 6, which are also visually similar.

\subsection{Colorectal laparoscopy}
The colorectal dataset consists of 9 colorectal laparoscopies recorded at the University Hospital of Heidelberg.
These 9 interventions are made up out of 6 proctocolectomies and 3 rectal resections.
While these interventions were recorded in the same manner as the dataset outlined in section \ref{sec:dataset}, the two datasets are disjunct.
Each of these videos was segmented into 8 phases (see table \ref{tab:phasesColo}) by the same surgical expert.
\begin{table}[htb]
\centering
\begin{tabular*}{\columnwidth}{l|l}
Phase ID & Explanation \\
\hline
1 & Team Time-Out\\
2 & Preparation and orientation at\\
  & abdomen\\
3 & Mobilization of colon\\
4 & Dissection of lymph nodes and blood\\
  & vessels\\
5 & Dissection and resection of rectum\\
6 & Preparation of anastomosis\\
7 & Placing stoma\\
8 & Finishing the operation\\
\hline
\end{tabular*}
\caption{Different phases in the colorectal dataset.}
\label{tab:phasesColo}
\end{table}

\begin{figure}[tb]
	\centering
	\subfigure[\textit{\NaiveNet}]{
	\label{colNoLSTM}
	\includegraphics[width=1.0\columnwidth]{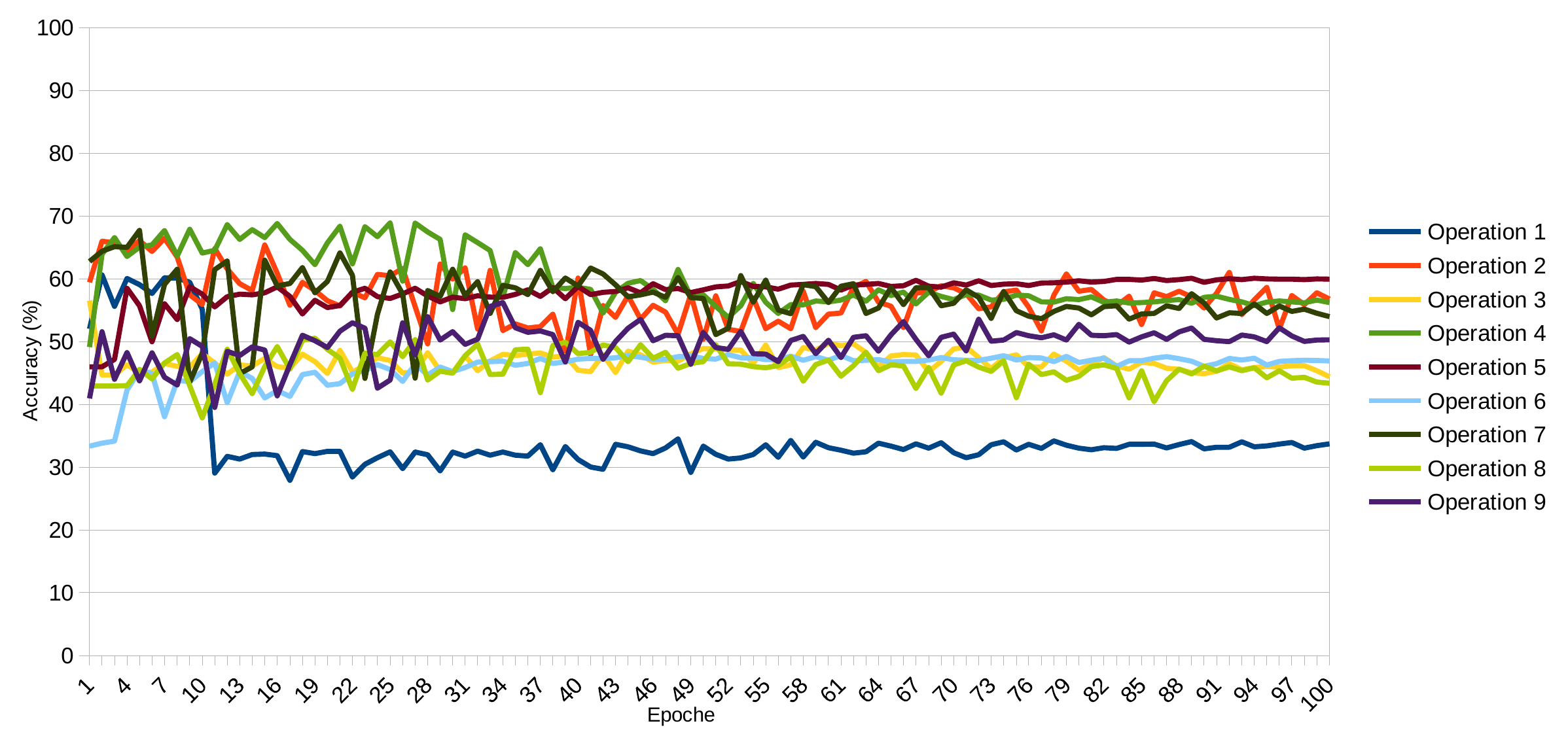}
	}
	\subfigure[\textit{\GRUNet}]{
	\label{colLSTM}
	\includegraphics[width=1.0\columnwidth]{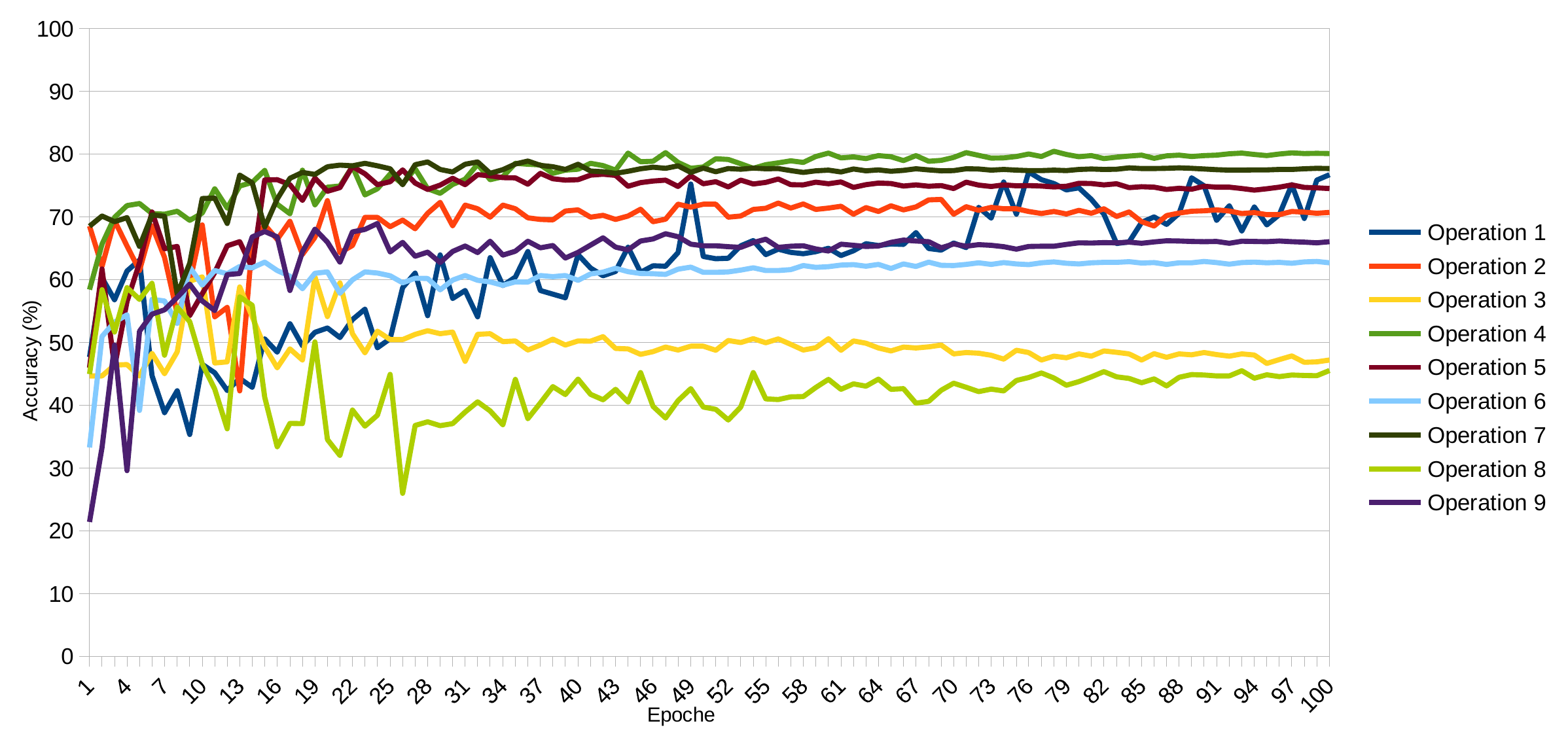}
	}
	\subfigure[\textit{\GRUNet} without pretraining]{
	\label{colNoLSTMNoPretrain}
	\includegraphics[width=1.0\columnwidth]{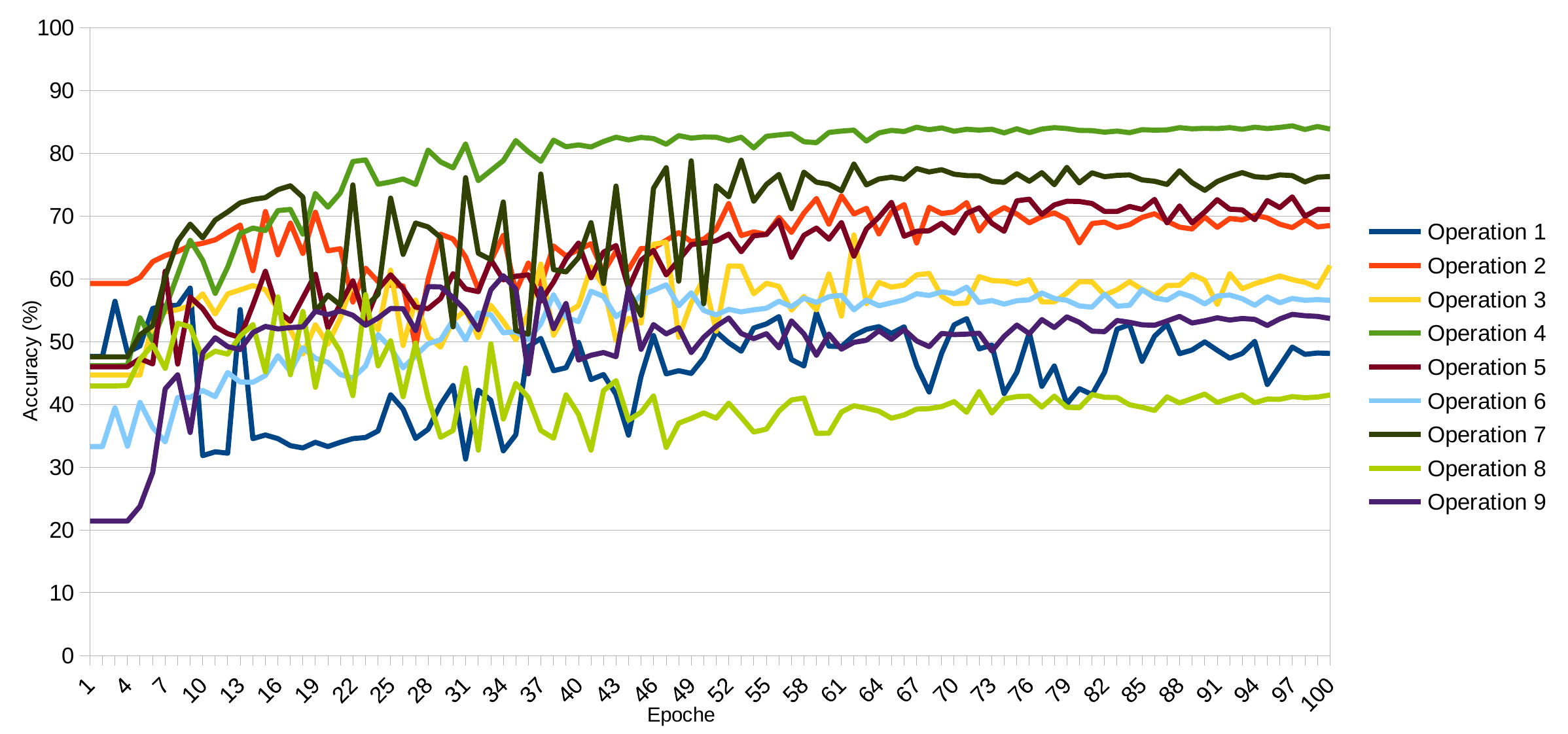}
	}
	\caption{Development of the accuracies of the phase detection for each operation and for each network during the leave-one-surgery-out evaluation on the colorectal dataset.}
	\label{fig:phaselstmcol}
\end{figure}

Similarly as to the previous section, we extracted one frame per second from the laparoscopic videos and resampled the frames to a resolution of $320\times 240$.
With this dataset, we then performed a leave-one-surgery-out evaluation for both \textit{\NaiveNet} and \textit{\GRUNet}.
For each test set, we trained for 100 epochs.
The same evaluation was also performed for a version of \textit{\GRUNet} with no pretrained weights.
The progression of the accuracies of each test run for each method can be found in figure \ref{fig:phaselstmcol}.
The graphs clearly show that even for this dataset, the GRU based methods achieve a higher accuracy than \textit{\NaiveNet}.
As seen in the previous section, the pretraining also boosts the classification performance on this dataset.

\begin{table}[htb]
\centering
\resizebox{\columnwidth}{!}{
\begin{tabular}{|l|l|l|l|}
\hline
& Precision & Recall & Accuracy \\
\hline
\textit{\NaiveNet} & 32.0\% $\pm$ 9.6\%& 29.7\% $\pm$ 8.5\% & 50.4\%\ $\pm$ 9.0\%\\
\hline
\textit{\GRUNet} & 68.2\% $\pm$ 15.0\%& 52.6\% $\pm$ 9.8\% & 67.2\%\ $\pm$ 13.1	\%\\
\hline
\textit{\GRUNet} without pretraining & 53.9\% $\pm$ 6.7\%& 43.6\% $\pm$ 11.2\% & 62.8\%\ $\pm$ 14.1\%\\
\hline
\end{tabular}
}
\caption{Comparison of the results of our proposed methods on the colorectal dataset from the University of Heidelberg.}
\label{tab:resultsCol}
\end{table}

This assumption is confirmed by table \ref{tab:resultsCol}.
A pretrained \textit{\GRUNet} achieves higher values for precision, recall and accuracy than \textit{\GRUNet} without pretraining and \textit{\NaiveNet}.

\begin{table}[hbt]
%\begin{table*}[t]
\centering
\resizebox{\columnwidth}{!}{
% \begin{tabular}{|l|l|l|l|l|l|l|l|l|}
% \hline
% & P1 & P2 & P3 & P4 & P5 & P6 & P7 & P8\\
% \hline
% Precission & 88.1\%$\pm$28.0\% & 72.9\%$\pm$24.1\% & 72.7\%$\pm$15.7\% & 58.7\%$\pm$43.9\% & 76.7\%$\pm$14.1\% & 57.7\%$\pm$31.0\% & 55.7\%$\pm$52.5\% & 62.9\%$\pm$45.1\%\\
% \hline
% Recall & 85.8\% $\pm$30.2\% & 67.0\%$\pm$33.3\% & 74.8\%$\pm$31.2\% & 9.3\%$\pm$17.1\% & 80.3\%$\pm$18.7\% & 37.0\%$\pm$37.0\% & 11.5\%$\pm$33.2\% & 51.3\%$\pm$42.3\%\\
% \hline
% Accuracy & 99.5\%$\pm$0.7\% & 97.8\%$\pm$1.4\% & 83.4\%$\pm$5.5\% & 91.4\%$\pm$5.8\% & 80.8\%$\pm$9.7\% & 88.2\%$\pm$10.2\% & 97.4\%$\pm$2.3\% & 96.8\%$\pm$3.5\%\\
% \hline
% \end{tabular}
\begin{tabular}{|l|l|l|l|}
\hline
& Precision & Recall & Accuracy\\
\hline
P1 & 88.1\%$\pm$28.0\% & 85.8\%$\pm$30.2\% & 99.5\%$\pm$0.7\%\\
\hline
P2 & 72.9\%$\pm$24.1\% & 67.0\%$\pm$33.3\% & 97.8\%$\pm$1.4\%\\
\hline
P3 & 72.7\%$\pm$15.7\% & 74.8\%$\pm$31.2\% & 83.4\%$\pm$5.5\%\\
\hline
P4 & 58.7\%$\pm$43.9\% & 9.3\%$\pm$17.1\% & 91.4\%$\pm$5.8\%\\
\hline
P5 & 76.7\%$\pm$14.1\% & 80.3\%$\pm$18.7\% & 80.8\%$\pm$9.7\%\\
\hline
P6 & 57.7\%$\pm$31.0\% & 37.0\%$\pm$37.0\% & 88.2\%$\pm$10.2\%\\
\hline
P7 & 55.7\%$\pm$52.5\% & 11.5\%$\pm$33.2\% & 97.4\%$\pm$2.3\%\\
\hline
P8 & 62.9\%$\pm$45.1\% & 51.3\%$\pm$42.3\% & 96.8\%$\pm$3.5\%\\
\hline
\end{tabular}
}
\caption{Performance of \textit{\GRUNet} on the colorectal dataset broken down into the different phases.}
\label{tab:results2Col}
\end{table}

The phase-wise performance of \textit{\GRUNet} is listed in table \ref{tab:results2Col}.
Phases 4 and 7 achieve the lowest performance.
Phase 4 is often so confused with phase 3, which precedes it and phase 5, which generally follows it.
Phase 7 is a rather short phase, meaning only a small number of examples were available for training and visually similar to phase 5 with which it is often confused.

\section{Discussion}
In this paper, we presented a method that allows us to train a CNN to differentiate between frames taken from the same video in a temporal context.
To train such a CNN, only the temporal order between two frames is required, which can be inferred from a given video.
Therefore no additional manual annotations by a surgical expert are required

Furthermore, we showed that such a network can be adapted to solve certain video segmentation tasks, in particular surgical phase detection.
We evaluated the method on two datasets: a publicly available dataset of annotated cholecystectomies and a dataset of annotated colorectal interventions.
The evaluation showed that on both datasets a GRU-based approach outperforms a plain feed-forward network.
A combination of the GRU-based approach and the pretrained model further increased performance, supporting our hypothesis that applying the pretraining method outlined in section \ref{sec:tcl} would be beneficial.

Our proposed method, which combines pretraining and a GRU, performs comparable to the state of the art on the public dataset, while the feedforward and the non-pretrained method perform significantly lower.
\textit{\GRUNet} outperforms the method of Dergachyova et al.\cite{Dergachyova2016} and the purely CNN-based EndoNet\cite{TwinandaSMMMP16}, which did not include temporal information. 
A second version of EndoNet incorporates temporal information using a hierarchical hidden markov model and thereby achieves a higher performance than \textit{\GRUNet}.
When comparing the performance of EndoNet and \textit{\GRUNet}, one has to take into consideration that EndoNet used 40 further annotated cholecystectomies for training.

A laparoscopic cholecystectomy is a very standardized and simple intervention.
Therefore, to show that our method can also be applied to longer, more complex laparoscopic interventions, we performed another evaluation on a dataset consisting of colorectal interventions, which are generally more complex in terms of involved anatomy, vessel resection and required level of surgical expertise.
The resulting performance was lower than on the cholecystectomy dataset.
This, in our opinion, can be attributed to the large variance in the dataset, which should be expected with long and complex interventions.
The order of certain phases varied partially between different interventions, e.g. in operation 7 phase 7 was not performed and in most operations, phase 3 was interrupted multiple times by other phases.
This can be partially attributed to the fact that the interventions were performed by different surgeons, as different surgeons have different preferences when it comes to the order of certain parts of the procedure.
The endoscopic optic and the tools used also varied between interventions.
This leads us to conclude that more examples, which mirror this variance, are required to increase performance.
Nevertheless, we were able to show that our pretrained CNN achieves a higher performance on this dataset than a randomly initialized CNN.
To improve the result of the phase segmentation, further post-processing steps, e.g. a sliding window or a hidden markov model, could be applied to the output of our CNN for smoothing.

In addition to surgical phase detection, the pretrained network could possible be used for other tasks in laparoscopy.
One application could be other segmentation tasks, such as action detection or event recognition. 
Furthermore, the of output of layer fc6 could be used as a reduced representation of a laparoscopic frame for allowing indexing of surgical videos.

\section*{Acknowledgements}
\label{acknowledgments}
The presented research was conducted in the project A01 within the setting of the SFB/Transregio 125 ``Cognition-Guided Surgery'' funded by the German Research Foundation. 
It is furthermore sponsored by the European Social Fund of the State Baden-Wuerttemberg and Heidelberg Medical School with a Physician-Scientist-Fellowship.
The authors also like to thank NVidia for the sponsored GeForce Titan X and also Ralf Stauder of the Technische Universit\"at M\"unchen (TUM) for providing the EndoVis Workflow dataset.
%\section*{References}

\bibliography{mybibfile}

\end{document}